\pdfoutput=1

\documentclass[11pt]{article}

\usepackage[preprint]{acl}

\usepackage{times}
\usepackage{latexsym}

\usepackage[T1]{fontenc}

\usepackage[utf8]{inputenc}

\usepackage{microtype}

\usepackage{inconsolata}

\usepackage{amsmath,amsfonts,bm}

\def\eqref#1{equation~\ref{#1}}

\def\1{\bm{1}}

\DeclareMathAlphabet{\mathsfit}{\encodingdefault}{\sfdefault}{m}{sl}
\SetMathAlphabet{\mathsfit}{bold}{\encodingdefault}{\sfdefault}{bx}{n}

\usepackage[utf8]{inputenc} 
\usepackage[T1]{fontenc}    

\usepackage{url}
\usepackage{microtype}
\usepackage{graphicx}
\usepackage{subfigure}
\usepackage{booktabs} 
\usepackage{amsfonts}       
\usepackage{nicefrac}       
\usepackage{microtype}      
\usepackage{xcolor}         
\usepackage{booktabs,tabularx,colortbl,multirow,array,makecell}

\usepackage[capitalize,noabbrev]{cleveref}
\usepackage{url}            
\usepackage{nicefrac}       
\usepackage{bm}

\usepackage{times}
\usepackage{latexsym}

\usepackage{amsmath}

\usepackage{algorithm}
\usepackage{algpseudocode}
\usepackage{graphicx}

\usepackage{bigstrut,bigdelim}
\usepackage{paralist}
\usepackage{diagbox}
\usepackage{wrapfig}
\usepackage{verbatim}

\usepackage{multicol}
\usepackage{multirow}
\usepackage{amssymb}
\usepackage{xcolor}
\usepackage{caption}
\usepackage{acronym}
\usepackage{xspace}
\usepackage{pifont}
\usepackage{lipsum}
\usepackage{enumitem}

\makeatletter
\DeclareRobustCommand\onedot{\futurelet\@let@token\@onedot}
\def\@onedot{\ifx\@let@token.\else.\null\fi\xspace}

\def\etc{\emph{etc}\onedot}

\makeatother

\makeatletter
\renewcommand{\paragraph}{%
  \@startsection{paragraph}{4}%
  {\z@}{0ex \@plus 0ex \@minus 0ex}{-1em}%
  {\hskip\parindent\normalfont\normalsize\bfseries}%
}
\makeatother

\makeatletter
\newcommand{\thickhline}{%
    \noalign {\ifnum 0=`}\fi \hrule height 1pt
    \futurelet \reserved@a \@xhline
}
\makeatother

\acrodef{nlp}[NLP]{natural language processing}
\acrodef{vqa}[VQA]{Visual Question Answering}
\acrodef{cs}[CS]{coherence scoring}
\acrodef{avsd}[AVSD]{coherence scoring}
\acrodef{roi}[RoI]{Region-of-Interest}
\acrodef{llm}[LLM]{large language model}
\acrodef{tom}[ToM]{theory-of-mind}
\acrodef{bce}[BCE]{Binary Cross Entropy}
\acrodef{mse}[MSE]{mean squared error}

\graphicspath{figures}

\title{Enhancing Common Ground Alignment and Negotiation through Theory of Mind Modeling}

\author{Shuwen Qiu$^1$, Mingdian Liu$^2$, Hengli Li$^{3,4}$, Song-Chun Zhu$^{4,5,6}$, Zilong Zheng$^{4}$ \\
$^1$ University of California, Los Angeles, $^2$ Iowa State University\\ 
        $^3$ Yuanpei College, Peking University\\
        $^4$ Beijing Institute for General Artificial Intelligence (BIGAI)\\
        $^5$ Institute for Artificial Intelligence, Peking University\\
        $^6$ Department of Automation, Tsinghua University\\
        s.qiu@ucla.edu, mingdian@iastate.edu \\
        lihengli@stu.pku.edu.cn, s.c.zhu@pku.edu.cn, zlzheng@bigai.ai\\}

\begin{document}
\maketitle
\begin{abstract}
Humans talk in daily conversations while aligning and negotiating the expressed meanings or common ground. Despite the impressive conversational abilities of the large generative language models, they do not consider the individual differences in contextual understanding in a shared situated environment. In this work, we propose MindDial, a novel conversational framework that can generate situated free-form responses with \ac{tom} modeling. We introduce an explicit mind module that can track the speaker's belief and the speaker's prediction of the listener's belief. Then the next response is generated to resolve the belief difference and take task-related action. Our framework is applied to both prompting and fine-tuning-based models, and is evaluated across scenarios involving both common ground alignment and negotiation. Experiments show that models with mind modeling can achieve higher task outcomes when aligning and negotiating common ground. The ablation study further validates the three-level belief design can aggregate information and improve task outcomes in both cooperative and negotiating settings.
\end{abstract}

\section{Introduction}
\begin{figure*}
    \captionsetup{type=figure}
    \includegraphics[width=\textwidth]{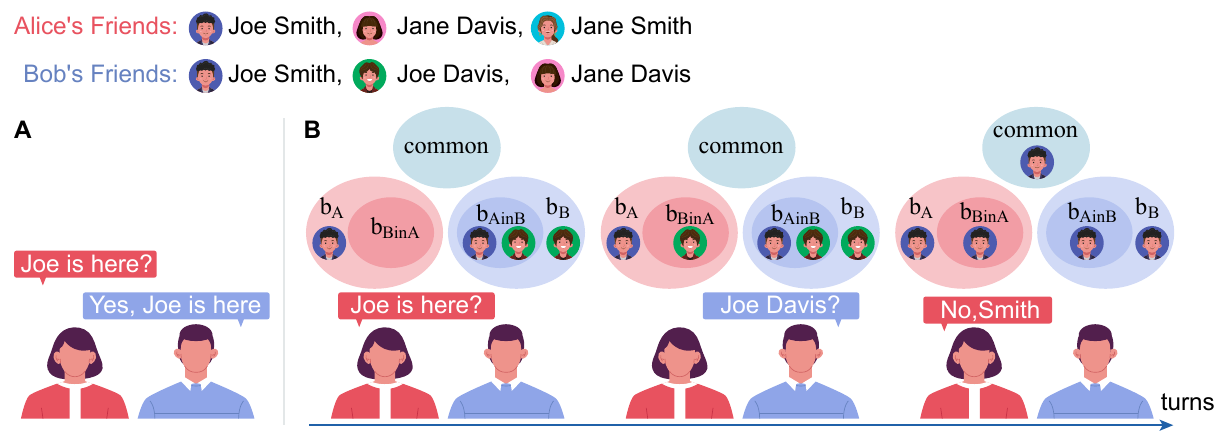}
    \captionof{figure}{\textbf{Left:} Single-turn question answering. \textbf{Right:} Multi-turn common ground alignment. Speakers will update their belief estimation based on context and generate the next response to reduce the belief differences.}
    \label{fig:teaser}
\end{figure*}

We align and negotiate our common ground every day in daily chit-chat~\cite{clark1986referring, bazerman2000negotiation}. In a common ground alignment scenario, agents are talking toward a joint goal, topics ranging from daily trivia to important multi-party meetings. In common ground negotiation situations, two parties resolve the differences in their beliefs, intents, or goals in a way that both find acceptable, such as item trading and discussing job offers~\cite{veinott1999video, beers2006common}. Though it seems easy between human conversations, it requires complicated social capabilities. Importantly, for all types of human communication including language, the relationship between the overt communicative act and common ground – of whatever type – is complementary. That is, as more can be assumed to be shared between communicator and recipient, less needs to be overtly expressed~\cite{tomasello2010origins}. Taking~\cref{fig:teaser}B as an example, when Bob asks about ``Joe Davis'', Alice will align the precise referents of the query by keeping ``Joe'' but correct ``Davis'' to ``Smith''. In this process, people need to realize what is shared and what needs to be further aligned or negotiated – which requires the understanding between points of view from their own and others' perspectives~\cite{blutner2000some, de2015higher} – the cognitive capability known as theory-of-mind (ToM).

The recent surge of large language models~(LLMs)~\cite{radford2019language,brown2020language} have dominated the natural language processing~(NLP) community for their prominent natural language generation performance. Although LLMs have shown their potential in ToM benchmarks~\cite{kosinski2023theory, ullman2023large, sileo2023mindgames, kim-etal-2023-fantom}, applying \ac{tom} for situated dialogue generation remains underexplored. In these situated tasks, agents' interactions are influenced by the environment, their shared experiences, and immediate goals.  The participants need to take into account not only the linguistic content but also factors such as the social context, prior knowledge, and each other's beliefs. Without \ac{tom}, the models can only provide the most possible response as a one-turn question-answering as shown in~\cref{fig:teaser}A. To enable LLMs to interact with people in a more socially realistic manner, it is essential to incorporate \ac{tom} for various forms of communications, such as aligning and negotiating common ground within dialogues~\cite{burleson2007constructivism, chiu-etal-2023-symbolic-planning, fu2023improving}. 

In this work, we introduce \textbf{MindDial}, a new dialogue framework designed to facilitate the alignment and negotiation of common ground in situated dialogues, incorporating \ac{tom} modeling. Inspired by the complementary role between common ground and communication, we design the two-step response generation. First, an explicit mind module estimates the speaker’s current perspective of the world (the first-order belief) and also helps speaker's estimate the other’s perspective of the world (the second-order belief)~\cite{grueneisen2015know, brauner2016recursive}. Then, the next response is aimed at resolving the belief difference. As shown in~\cref{fig:teaser}B, Alice says ``No, Smith'' when her first-order belief $b_A$ (``Joe Smith'') does not equal to her second-order belief $b_{BinA}$  (``Joe Davis'') .

In sum, we consider our contributions as three-fold:

i) We design a framework incorporating an explicit mind estimation module that tracks the first-order and second-order beliefs. Resolving the belief difference between the two will support the next response generation. 

ii) We explore two types of response generators -- fine-tuning and prompting-based models. The experiments show that our framework can successfully improve model performance in both groups. 

iii) We test our framework on both aligning and negotiating settings. The evaluation results and user study validate that our framework can improve both the cooperation and negotiation abilities of the LLM agents. We ablate each level of the beliefs and find both first and second-order contribute to the final results. 

\section{Related work}
\paragraph{Theory-of-Mind (ToM)}
ToM is a crucial capability for human social interactions developed in early life~\cite{kovacs2010social,richardson2018development}. In literature, early works model belief update through time in sequential games with partially observable Markov decision process (POMDP)~\cite{baker2011bayesian, de2013much, vogel2013emergence, doshi2010modeling,han2018learning}. One agent’s belief update is based on the estimate of others’ current beliefs, resulting in an infinite recursion. However, in real life, studies have shown that humans could go no deeper than two levels of recursion~\cite{camerer2004cognitive}. Therefore, works~\cite{fan2021learning} began the efforts to end the recursion when their beliefs merge into the “common mind”. 

Modeling the belief of others has been extensively studied in symbolic-like environments~\cite{wunder2011using,rabinowitz2018machine,kleiman2016coordinate,ho2016feature}, where agents need to incorporate or compete for a goal. Efforts to measure models’ ability to recognize false beliefs and perspective-taking also emerge in robotics~\cite{yuan2020joint,milliez2014framework}, computer vision~\cite{eysenbach2016mistaken, fan2021learning}, and natural language processing~\cite{qiu-etal-2022-towards, nematzadeh-etal-2018-evaluating} using the Sally-Anne test~\cite{baron1985does}. Different variants of the Sally-Anne test and ToM benchmarks are also proposed to test the ToM of large language models~\cite{kosinski2023theory, ullman2023large, sileo2023mindgames, kim-etal-2023-fantom}. It is also shown that augmenting the model with external mind modules can help improve the performance of tasks involving intensive belief exchange and rich social interaction scenarios~\cite{fan2021learning, qiu-etal-2022-towards, li-etal-2023-theory, chiu-etal-2023-symbolic-planning}. In this work, we explore \ac{tom} modeling can enhance the quality and efficiency of the response generation in both cooperative and semi-cooperative dialogue tasks.

\paragraph{Common ground alignment and negotiation}
 In a cooperative dialogue task, to guarantee that the communication takes the least cost meanwhile providing the most informative messages,  previous works proposed multiple methods to align the common ground between agents~\cite{bohn2019integrating, anderson2021tell}. Specifically for dialogue tasks, datasets have been collected to provide golden utterances when people try to align the common ground with each other based on structured knowledge~\cite{he-etal-2017-learning}, in partially observable cooperative tasks~\cite{bara-etal-2021-mindcraft,kim-etal-2019-codraw}, in multimodal and continuous environment~\cite{haber-etal-2019-photobook, udagawa-aizawa-2021-maintaining}. Frameworks have been adopted to model and predict the aligning dynamics using GNN, RNN, transformers, and LLMs~\cite{he-etal-2017-learning, udagawa-aizawa-2021-maintaining, fischer2023reflective, zhang2023building, zhou2023far}. The inferred common ground is also used to generate more interesting and engaging conversations for the dialogue agents~\cite{zhou-etal-2022-reflect}.  

Negotiation is treated as a semi-cooperative task since agents can have different goals but need to agree on the same decision~\cite{lewis-etal-2017-deal}. It requires complex social skills and strategies like offering proposals and accepting or making counter-offers~\cite{yamaguchi-etal-2021-dialogue}. To improve the negotiating abilities of the dialogue systems, datasets of open-domain human negotiation corpus have been introduced in embodied environment~\cite{devault2015toward}, daily items split~\cite{lewis-etal-2017-deal, chawla-etal-2021-casino}, buy and sell~\cite{he-etal-2018-decoupling}, job offer negotiation~\cite{yamaguchi-etal-2021-dialogue}. Modeling begins with game theory and action selection~\cite{nash1950bargaining, baarslag2013evaluating}. For open-domain generation, methods have been designed to help the model plan ahead~\cite{lewis-etal-2017-deal, iwasa2018prediction}, give feedback about the current conversation~\cite{zhou-etal-2019-dynamic, fu2023improving}, detect negotiation breakdowns~\cite{yamaguchi-etal-2021-dialogue}.
 
\begin{figure*}[t!]
    \centering
    \small
    \begin{tabular}{c}
    \includegraphics[width=0.9\linewidth]{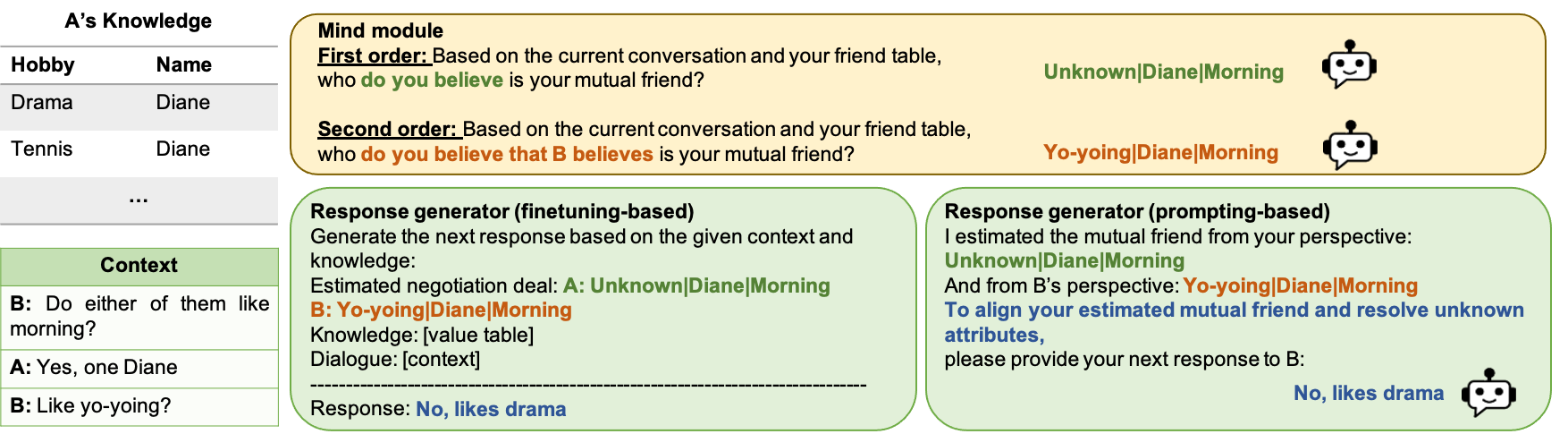} \\ \midrule
    \includegraphics[width=0.9\linewidth]{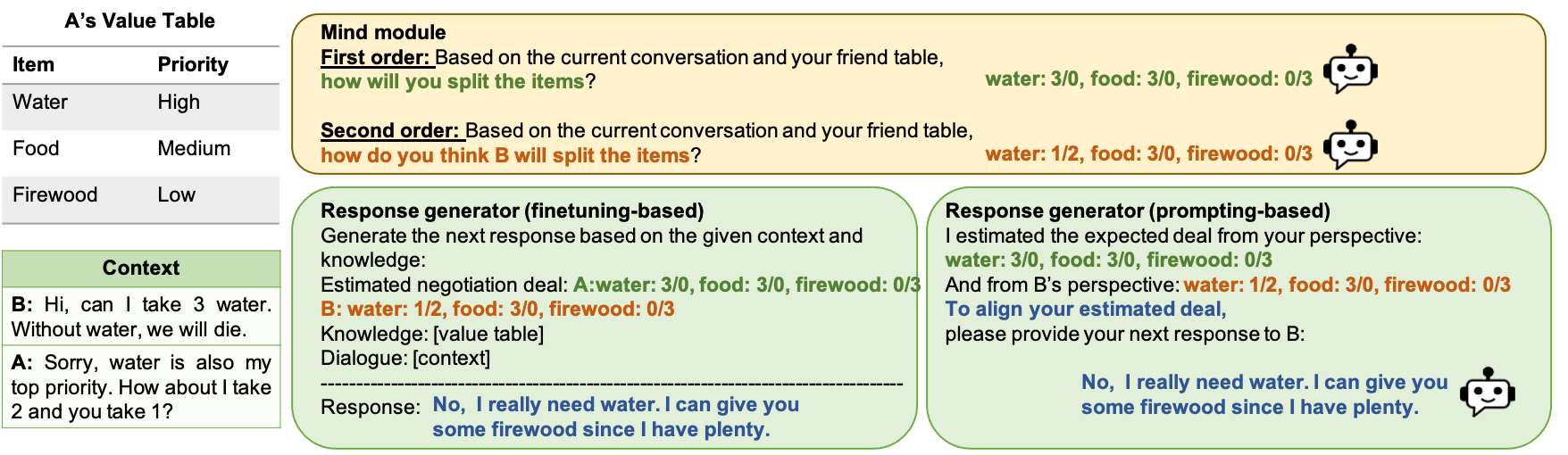}
    \end{tabular}

    \caption{\textbf{Cases of ToM reasoning in MindDial.} Top: an \textit{alignment} task from MutualFriend. Bottom: A \textit{negotiation} task from CaSiNo. For each task, we first reason over the first- and second-order ToM beliefs of the conversational partner. Then we generate corresponding utterances wrt. the ToM estimation.}
    \label{fig:mf_model}
\end{figure*}

\section{Task and Framework}
\subsection{Tasks}
The situated dialogue corpus can be denoted as $\mathcal{D}=\{(U_n, K_n^p,  y_n)\}_{n=1}^N$, where $U_n=(u_{n, 1}, ..., u_{n, T})$ represents the dialogue history and $T$ is the number of turns. $K_n^p=(k_{n,1}, ... k_{n,I})$ is for their knowledge base, where $I$ is the number of knowledge passages. $p \in {A, B}$ represents the two agents. We assume the current speaker is A, and $p$ will be dropped for the following formulation. $y_n$ is A's next response or its action to achieve the task goals. 
\paragraph{Alignment}
In the common ground alignment scenarios, we use the MutualFriend task~\cite{he2017learning} shown in~\cref{fig:mf_model}. $K$ denotes the private friend lists that two agents observe, and there is only one friend shared in their lists. The agents need to merge their estimation of the mutual friend through chat and finally finish the task goal by taking the action to select $k_i \in K$ as their mutual friend. The alignment is successful when their selections are the same.
\paragraph{Negotiation}
In the common ground negotiation scenarios, we use the CaSiNo task~\cite{chawla-etal-2021-casino} shown in~\cref{fig:mf_model}(Bottom). Two agents are planning a camp trip and need to distribute the uneven number of items. Based on their individual priority of the items $K$, they need to decide on the final item split agreement to maximize their gain of valuable items. At the end of the conversation, one agent proposes the item split deal while the other agent decides to accept or reject this deal. The negotiation is successful when the deal is accepted. 

\subsection{MindDial}
The overall pipeline of our framework is shown in~\cref{fig:mf_model}. At the first stage, given the context history and private knowledge, the mind module $f$ estimates the first and second-order beliefs over their solutions $b_A, b_{BinA} = f(U, K)$. The first-order belief represents $A$'s estimation of the mutual friend or split deal. The second-order belief refers to $A$'s understanding of $B$'s estimation regarding the mutual friend or split deal. We choose to prompt the LLMs for $b_A$ and $b_{BinA}$ due to their ability to adapt flexibly in open-domain corpora. Therefore, the mind module can be applied to other situated dialogues when the knowledge base and beliefs are well-defined. 

Then the response generator $h$ generates the next utterance based on the dialogue history, its private knowledge, and the intention to align the first and second-order beliefs: $\tilde{y} = h(U, K, b_A, b_{BinA})$. We apply two methods to the response generator to activate their ability to resolve the belief difference in $b_A$ and $b_{BinA}$: embedding this ability into LLM by finetuning and explicitly triggering this ability of LLM by prompting.
\paragraph{Finetuning-based} For finetuning-based models, we prepare a small dataset in the format of \{$y$, $U$, $K$, $b_A$, $b_{BinA}$\}, where $y$ is intended to resolve the gap between $b_A$ and $b_{BinA}$. Different parts of model inputs are concatenated together with their corresponding tags as [Estimated belief], [Knowledge], and [Dialogue] shown in~\cref{fig:mf_model}. The models are trained to regress the next response $y$. 
\paragraph{Prompting-based} For prompting-based models, we directly ask the generator to generate the next response in order to resolve the difference and unknown values between $b_A$ and $b_{BinA}$. The format follows as “I estimated mutual friend/deal from your perspective: $b_A$ and from B’s perspective: $b_{BinA}$. To align $b_A$ and $b_{BinA}$, please provide your next response to B:”. 

\section{Experiments}
\paragraph{Dataset}
To provide a reasonable quantitative measure of belief dynamics in the dialogue, the expected dataset should contain rich belief exchanges. Meanwhile, the belief exchange and the final solution can be easily labeled. Therefore, we choose two representative settings to evaluate our framework.  \textbf{MutualFriend}~\cite{he2017learning}: we consider it as an alignment dialogue scenario. In the MutualFriend task, each agent has a private knowledge base including a list of friends and their attributes like name, school, \etc. There is a shared friend that both agents have and they need to chat with each other to find this mutual friend. We only keep the successful dialogues and the final data split for train/val/test is 7257/878/900. Each dialogue in the training set contains a maximum of 53 turns and each turn with a maximum length of 29. 
\textbf{CaSiNo}~\cite{chawla-etal-2021-casino}: we consider it as the negotiating scenario. In the CaSiNo task, two agents need to split camping packages: 3 water, 3 firewood, and 3 food. Each of these items will be of either High, Medium, or Low value to each agent. The agents need to negotiate the distribution of the items through chat to maximize their final points calculated based on the number of items they get and the items’ corresponding values. The data split for train/val/test is 900/30/100. Each dialogue in the training set contains a maximum of 27 turns and each turn with a maximum length of 106. 

\textbf{Mind modules}\quad{} To serve as a mind module in this task, the model is expected to understand long conversation contexts and the concept of first and second-order beliefs. We choose LLaMA-2-7B-chat, LLaMA-2-13B-chat~\cite{touvron2023LLaMA}\footnote{https://github.com/facebookresearch/LLaMA-recipes/tree/main},  GPT-3.5, and GPT-4\footnote{gpt-3.5-turbo-1106, gpt-4-1106-preview} as our mind reasoner for their potentials in \ac{tom} benchmarks and the flexible abilities of mind reasoning in open-domain dialogues. 

\begin{table*}[t!]
\centering
\small
\resizebox{\textwidth}{!}{
\begin{tabular}{llccc|llccc}
    \toprule

    \textbf{Models} & \textbf{Mind level} & \textbf{$C$} & \textbf{$T$} & \textbf{$C_T$} & \textbf{Models} & \textbf{Mind level} & \textbf{$C$} & \textbf{$T$} & \textbf{$C_T$}\\
    \midrule
\multirow{4}{*}{LLaMA-7B-ft}&w/o mind & 24.67 & 9.09 & 2.71 & \multirow{4}{*}{LLaMA-13B-ft}&w/o mind & 36.33 & 6.64 & \textbf{5.47} \\

&$b_A$ & 28.33 & 7.92 & \textbf{3.58} &  &$b_A$ & 42.00 & 8.66 & 4.85 \\
&$b_{BinA}$ & \textbf{29.33} & 8.33 & 3.52 & &$b_{BinA}$ & 39.33 & 7.70 & 5.11 \\
&$b_A$+$b_{BinA}$ & 28.33 & 8.87 & 3.20 &  &$b_A$+$b_{BinA}$ & \textbf{44.67} & 8.85 & 5.05 \\
    \midrule

\multirow{4}{*}{GPT-3.5}&w/o mind & 10.67 & 5.74 & 1.86 & \multirow{4}{*}{GPT-4}&w/o mind & 75.00 & 9.72 & 7.71 \\

&$b_A$ & 18.33 & 5.91 & 3.10 & &$b_A$ & 75.00 & 9.41 & 7.97 \\
&$b_{BinA}$ & 12.33 & 5.91 & 2.09 & &$b_{BinA}$ & 69.67 & 8.84 & 7.88 \\
&$b_A$+$b_{BinA}$ & \textbf{24.33} & 6.04 & \textbf{4.03} & &$b_A$+$b_{BinA}$ & \textbf{76.00} & 8.88 & \textbf{8.56} \\
    \bottomrule
    \end{tabular}}
\caption{\textbf{MutualFriend: results with different mind settings}. Settings without mind reasoning are marked as w/o mind. Settings considering only the first-order are marked as $b_A$, with only the second-order are $b_{BinA}$, with both are $b_A$+$b_{BinA}$.}
\label{tab:mf_ablate}
\end{table*}
 
\textbf{Response generators}\quad{} We adopt the same four models in the mind modules as our response generators. We divide the models into two groups: finetuning and prompting-based. For the finetuning group, we first finetune LLaMA-2-7B-chat and LLaMA-2-13B-chat to generate the next response with the raw training dialogues. Then, we sample 3\% of the training data and predict the first and second-order beliefs at each turn using the mind module, which are put into the dialogue context as additional information input to finetune the model again. We choose to combine only a small portion of training data input with beliefs to reduce the API query cost. We also vary the portion to 1\%, 3\%, and 5\%. The sample size does not significantly influence the model performance (See Appendix~\ref{sec:sample}). For GPT-3.5 and GPT-4, we use prompts to regulate the conversation. For finetuning-based models, the models are trained on two A6000 GPUs for one epoch with an initial learning rate of 1e-4. The batch size is set to 64. For prompting-based models, we use the OpenAI API for experiments.

\subsection{Evaluation and results}
For evaluation, we focus on three main questions: 
\begin{itemize}[leftmargin=*, noitemsep, topsep=0pt]
    \item \textbf{Question 1:} Can mind reasoning improve the common ground alignment and negotiation results?
    \item \textbf{Question 2:} Which level of beliefs contributes to the performance gain?
    \item \textbf{Question 3:} What is the relation between belief estimation accuracy and conversation outcomes? 
\end{itemize}


\begin{table*}[t!]
\centering
\small
\resizebox{\textwidth}{!}{
\begin{tabular}{ll|cccccc}
    \toprule
    \textbf{Models} &\textbf{Mind level}& \textbf{Score-all} & \textbf{Sum} & \textbf{Agreed \%} & \textbf{Pareto} & \textbf{Score-agreed} & \textbf{Sum}\\
    \midrule
\multirow{4}{*}{LLaMA-7B-ft}&w/o mind & 8.10 vs 7.18 & 15.28 & 24.00 & 12.00 & 18.33 vs 14.50 & 32.83 \\
&$b_A$ & 12.94 vs 13.48 & \textbf{26.42}&\textbf{68.00}&20.00& 16.68 vs 17.47 & 34.15\\
&$b_{BinA}$ & 11.76 vs 13.36& 25.12&56.00&24.00& 16.54 vs 19.29 &\textbf{35.83}\\
&$b_A$+$b_{BinA}$ & 12.96 vs 12.70 & 25.66 & 62.00 & \textbf{26.00} & 17.84 vs 17.42 & 35.26\\
    \midrule
\multirow{4}{*}{LLaMA-13B-ft}&w/o mind & 15.38 vs 12.68 & 28.06 & 70.00 & 24.00 & 19.83 vs 15.97 & 35.80 \\
&$b_A$ &18.02  vs16.14 &34.16 &92.00&38.00 & 19.15 vs 17.11 &36.26\\
&$b_{BinA}$ & 17.02 vs14.50 &31.52 &82.00&30.00& 19.66 vs 16.59 &36.25\\
&$b_A$+$b_{BinA}$& 17.36 vs 17.32 & \textbf{34.68} & \textbf{92.00} & \textbf{40.00} & 18.43 vs 18.39 & \textbf{36.82} \\
\midrule
\multirow{4}{*}{GPT-3.5}&w/o mind & 15.00 vs 14.26 & 29.26 & 80.00 & 18.00 & 17.38 vs 16.57 & 33.95\\
&$b_A$ & 16.10 vs 17.08 & 33.18 & 90.00 & 22.00 & 17.22 vs 18.42 & 35.64\\
&$b_{BinA}$ & 16.72 vs 16.86 & \textbf{33.58} & \textbf{92.00} & 22.00 & 17.74 vs 17.89 & 35.63\\
&$b_A$+$b_{BinA}$ & 17.08 vs 15.18 & 32.26 & 86.00 & \textbf{26.00} & 19.05 vs 16.72 & \textbf{35.77}\\
\midrule
\multirow{4}{*}{GPT-4}&w/o mind & 16.84 vs 16.90 & 33.74 & 94.00 & 8.00 & 17.60 vs 17.66 & 35.26 \\
&$b_A$ & 16.72 vs 16.50 & 33.22 & 90.00 & 14.00 & 18.02 vs 17.78 & 35.80 \\
&$b_{BinA}$ & 17.40 vs 16.56 & 33.96 & 92.00 & 12.00 & 18.17 vs 17.39 & 35.56 \\
&$b_A$+$b_{BinA}$ & 17.54 vs 17.46 & \textbf{35.00} & \textbf{96.00} & \textbf{20.00} & 18.06 vs 17.98 & \textbf{36.04} \\
    \bottomrule
    \end{tabular}}
\caption{\textbf{CaSiNo: results with different mind settings}. Settings without mind reasoning are marked as w/o mind. Settings considering only the first-order are as $b_A$, with only the second-order are $b_{BinA}$, with both are $b_A$+$b_{BinA}$.}
\label{tab:casino_ablate}
\end{table*}

\paragraph{MutualFriend evaluation metrics}
We adopt the same metrics in~\citet{he2017learning}:
\begin{itemize}[leftmargin=*, noitemsep, topsep=0pt]
    \item Success rate ($C$): how many dialogues where the two agents select the true mutual friend.
    \item Conversation turns ($T$): the number of turns the agents take before the end of the conversation
    \item Success rate per turn ($C_T$): how efficient the conversation is. We divide the overall success rate by the conversation turns. 
\end{itemize}

\paragraph{CaSiNo evaluation metrics}
We follow the procedure in~\citet{lewis-etal-2017-deal}:
\begin{itemize}[leftmargin=*, noitemsep, topsep=0pt]
    \item Score-all: the average negotiation scores. The points each agent scores are the number of items times the item’s corresponding values. High priority is a value of 5. Medium is 4. Low is 3. If the deal is rejected or the negotiation exceeds the maximum turn, both agents receive 5 points. Since the best outcome should be a win-win situation, we also report the sum over the points of the two agents to compare the overall performance gain.
    \item Agreed \%: the agreement of the deal. A deal is agreed when the agents agree on the proposed deal and the proposal does not exceed the total number of items the agents can distribute.
    \item Pareto: whether the deal is Pareto Optimal. A solution is Pareto Optimal if neither agent’s score can be improved without lowering the other’s score.
    \item Score-agreed: the average negotiation scores in agreed deals. 
\end{itemize}

\subsubsection{Observation I: Mind reasoning improves conversation outcomes}~\label{sec:quan}
First, our experiments compare models’ performance without and with mind reasoning. In the cooperative scenario in~\cref{tab:mf_ablate}, comparing model+w/o mind rows with models, we can see that combining mind modules can significantly improve the alignment success rate in both finetuning and prompting-based models. Among them, GPT-4 performs the best, following LLaMA and GPT-3.5. As for efficiency, models with mind reasoning exhibit higher per-turn success. However, for LLaMA13b, we notice a longer conversation length, therefore, the efficiency drops below the base model. This suggests that while incorporating belief estimation can elevate success rates, it may not necessarily enhance efficiency if acquiring additional information is needed to establish common ground. More comparison results and discussion can be found in Appendix~\ref{sec:more_mf_res}.

In the negotiation scenario, as referenced in~\cref{tab:casino_ablate}, agents utilizing mind reasoning capabilities tend to achieve higher individual scores. Additionally, the collective points of both parties are increased. These agents also are more likely to reach agreements and achieve Pareto Optimal outcomes, suggesting a more strategic distribution of items. When comparing the points scored and the agreement rates across different models, GPT variants and LLaMA-13B display similar performances except that  LLaMA-7B falls behind. Notably, LLaMA-13B achieves the highest Pareto Optimal scores, surpassing GPT-4. This may be attributed to GPT-4's tendency to favor equitable item distribution, often resulting in a split like 1 and 1, with another item left unclaimed by either party (examples can be found in Appendix~\ref{sec:qual_casino}). 
\begin{figure*}[t!]
\begin{minipage}{0.48\linewidth}
    \centering
    \small
    \begin{tabular}{l|ccc}
        \toprule
        \textbf{Models} & \textbf{Belief} & \textbf{Precision} & \textbf{F1}\\
        \toprule
        \multirow{2}{*}{LLaMA-7B}&$b_A$& 33.00&33.00\\
         &$b_{BinA}$&30.00&30.00 \\
        \midrule
        \multirow{2}{*}{LLaMA-13B}&$b_A$&36.00&34.00 \\
         &$b_{BinA}$&38.00&33.00 \\
         \midrule
        \multirow{2}{*}{GPT-3.5}&$b_A$&62.00&62.00 \\
         &$b_{BinA}$&70.00&67.00 \\
         \midrule
        \multirow{2}{*}{GPT-4}&$b_A$&77.00&77.00 \\
         &$b_{BinA}$&76.00&76.00 \\
        \bottomrule
        \end{tabular}
    \captionof{table}{\textbf{Belief prediction}. The precision and F1 when different models predict the first ($b_A$) and second-order ($b_{BinA}$) beliefs.}
    \label{tab:delta}

\end{minipage}%
\hfill%
\begin{minipage}{0.5\linewidth}
    \centering
    \includegraphics[scale=0.25]{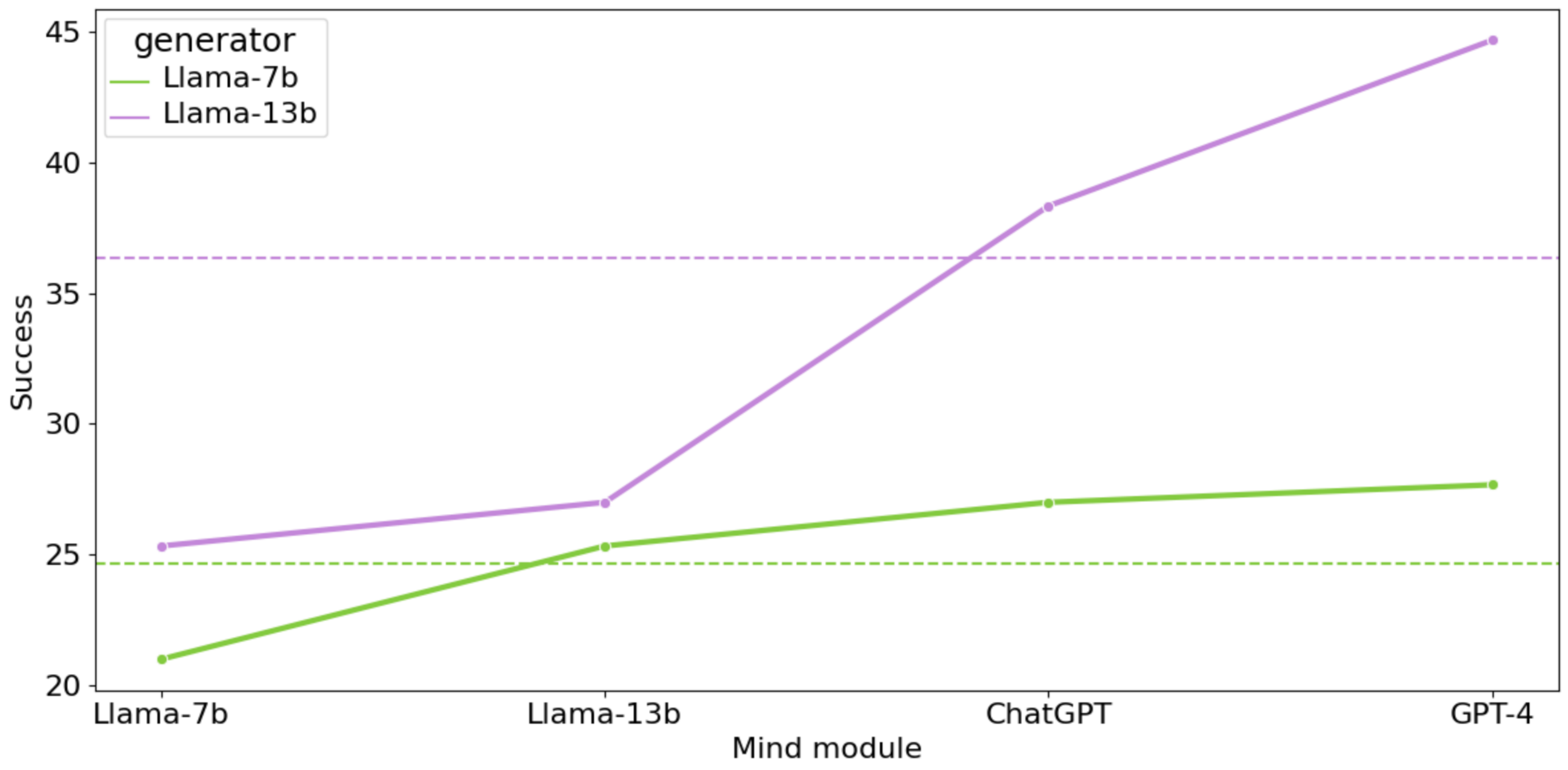}
    \captionof{figure}{The task success rate when the response generators are paired with different models as the mind modules. The X-axis marks the model name of the mind modules. The Y-axis shows the success rate. Different colors represent different models as the response generators. }
    \label{fig:mindvssuccess}
\end{minipage}%
\end{figure*}

\subsubsection{Observation II: Both two levels of belief contribute to the performance gain}
Next, we assess the impact of varying belief estimation levels in the mind modules on model performance, as shown in~\cref{tab:mf_ablate} and~\cref{tab:casino_ablate}. First, It is evident that integrating any level of belief estimation leads to performance enhancements compared with the w/o mind baseline, indicating both first and second-order beliefs contribute to the response generation process. Within mind settings, in alignment scenarios, models underscoring the belief differences usually outperform others with single-order belief estimation in LLaMA-13B, GPT-3.5, and GPT-4. In negotiating settings, we first notice that the score-all strongly correlates with the agreed rate, and there is no consistent pattern. Examining Pareto Optimal outcomes, models aggregating both $b_A$ and $b_{BinA}$ tend to distribute items more effectively, resulting in higher Pareto Optimal scores. Similarly, in score-agreed, models combining both two levels of beliefs perform better. 

We also notice some fluctuations in the results, for example, LLaMA-7B with only $b_{BinA}$ achieves better results. We reckon that complex and intertwined effects can be exerted when 1) the model is bottlenecked by its context understanding and generation abilities and 2) one or both levels of the belief estimations are not accurate. In general, models need to take into account their own beliefs and also the beliefs of others. Focusing on resolving the differences between them can improve the common ground alignment accuracy and negotiation optimality.

\paragraph{Robustness to prompts} In our experimental investigations (prompt templates are supplemented in Appendix~\ref{sec:mf_prompt} and \ref{sec:casino_prompt}), we found that the performance of belief prediction remains robust when prompts are structured to inquire about the current speaker's solution and their estimation of the other speaker's solution. 
Our comparison results in \cref{tab:mf_ablate,tab:casino_ablate} suggest that the task of one-hop prediction, encompassing beliefs and intentions, poses a minimal challenge for most LLMs. For instance, LLaMA-13B exhibits performance akin to GPT-3.5. Consequently, we assert that the primary challenge lies in advancing higher-level ToM inferences within these models.

\begin{figure*}[t!]
    \centering
    \small
    \includegraphics[width=0.9\linewidth]{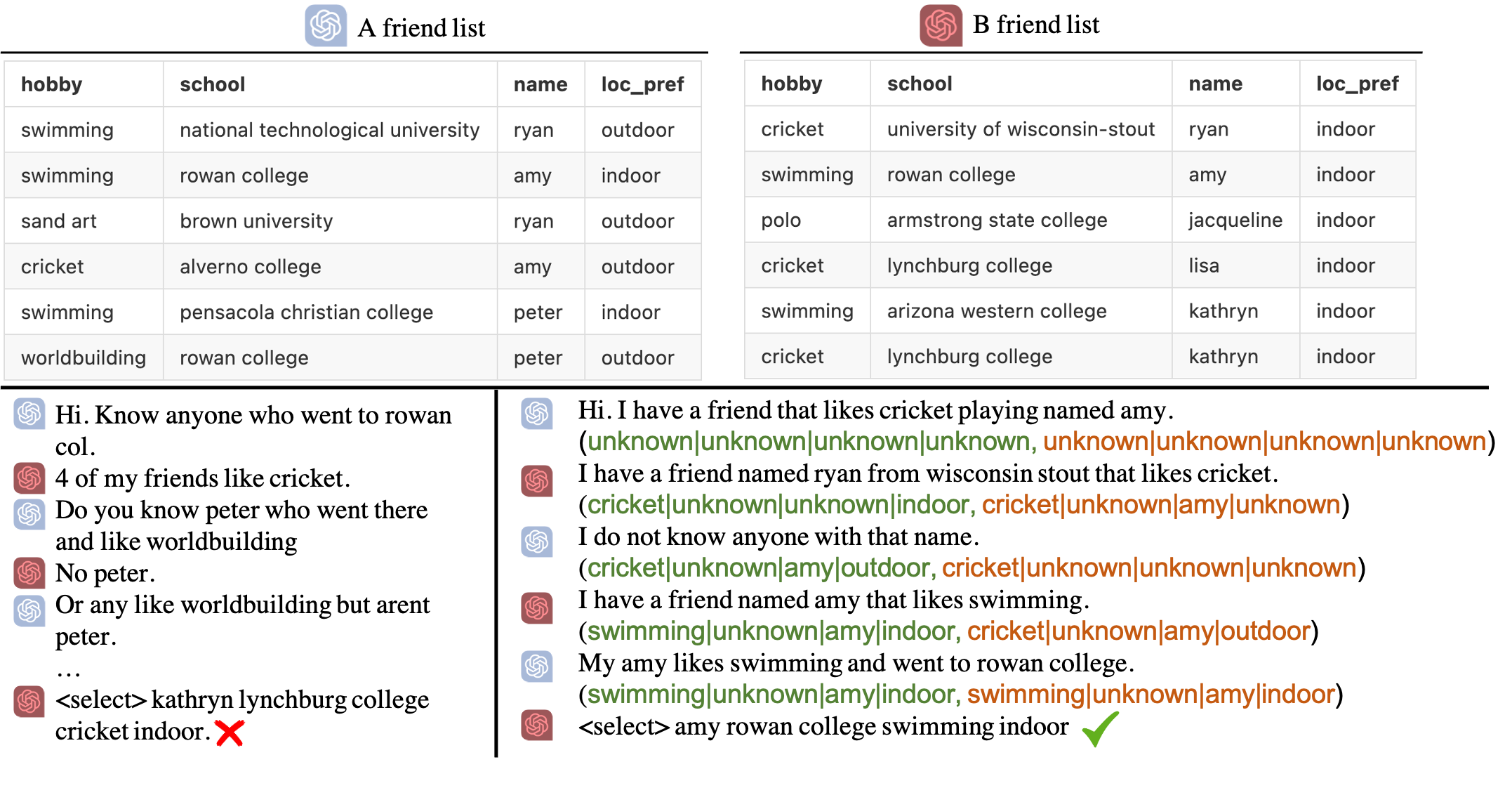}
    \caption{\textbf{Qualitative comparisons between dialogue generation models} without (at left) and with mind modeling (at right) when agents A and B are figuring out their mutual friend. }
    \label{fig:qual}
\end{figure*}
\subsubsection{Observation III: Belief estimation accuracy positively correlates with the alignment success}
To more convincingly validate that incorporating the mind reasoning module enhances the models’ task performance, we assessed the belief estimation accuracy when different models serve as the mind module in the MutualFriend task using LLaMA-7B, LLaMA-13B, GPT-3.5, and GPT-4. Subsequently, we examined how this accuracy correlates with task success when the four models function as response generators separately, paired with these four mind modules. Here, we demonstrate the relation between the belief estimation accuracy and the dialogue outcomes with MutualFriend task due to its clearly defined belief dynamics. Therefore, the first and second-order beliefs can be easily annotated using predefined rules. The detailed labeling process is included in Appendix~\ref{sec:belief label}. \cref{tab:delta} shows the precision and F1 scores for predicting the current speaker’s estimation of mutual friend given the current dialogue history $b_A$, and its estimation of the other speaker’s estimation $b_{BinA}$. The line plot in \cref{fig:mindvssuccess} illustrates the corresponding success rates when a response generator is equipped with different mind modules. 

Combining the models’ precisions of the belief prediction with the success rates when they serve as the mind modules, we can observe that 1) The success rates increase when the models with higher belief prediction accuracy are served as the mind modules. This trend underscores that the effectiveness of the response generators is closely linked to the mind reasoning capabilities of the respective mind modules; 2) Comparing the growth magnitude of LLaMA-7B and 13b, we can see that LLaMA-7B reaches a flat stage and increases slowly. This suggests that the magnitude of the success rate improvement is bounded by the model’s mind-reasoning abilities; 3) The horizontal lines mark the task success rate when the response generators are not augmented with the mind modules. Augmenting models with weaker mind modules can detrimentally impact outcomes due to inaccurate belief predictions and inadequate dialogue reasoning, such as the situation when LLaMA-7B is paired with LLaMA-7B and LLaMA-13B is paired with LLaMA-7B and 13b. 

\paragraph{Summarization \textit{vs.} Reasoning} It is worth noting that both the first- and second-order belief estimation goes beyond summarization from the last utterance.  We carefully annotate part of the beliefs in the dialogue and report the second-order belief prediction accuracy in \cref{tab:delta}, which shows that the LLM can predict the second-order beliefs fairly
well.

\subsubsection{Human Evaluation}
We ask 16 college-level students to play MutualFriend and CaSiNo game with our model. Each subject is randomly assigned 4 samples. S/He chooses one sample to play with the agent w/o mind modules and the other one to play with the agent w/ mind modules. A pair-wise comparison is made between the game outcome when human subjects play with models without and with mind reasoning. In addition, after the game ends, the subjects rate their game partner regarding their cooperativeness (whether the agent is cooperative during the game) and informativeness (whether the agents provide informative responses) from 0 to 10 in the alignment setting; rate regarding their negotiation skills (whether the agent is a good negotiator) and whether they are satisfied with the final deal in the negotiation setting. In addition, we also record their overall enjoyment when playing with the agents in both settings. 

From Table~\ref{tab:human}, we can observe that our model with mind modules can achieve higher outcomes in both MutualFriend and CaSiNo games and the subjects tend to enjoy more in the process. In the cooperative setting, agents without and with mind achieves similar cooperativeness and informativeness rates. However, in the negotiation setting, agents with mind reasoning are shown to be more skillful and can achieve more satisfactory deals. Please refer to Appendix~\ref{sec:interface} for interface details.
\subsection{Case Study}
We demonstrate one MutualFriend example to visualize the difference between LLaMA-7B with and without mind reasoning. Examples of other models and CaSiNo scenarios can be found in the Appendix. As shown in~\cref{fig:qual}, the topics between agents without mind reasoning can diverge quickly. For example, when A asks about ``Rowan College’’, B responds with ``cricket’’ which is unrelated to it. In contrast, for dialogues between agents with step-wise mind reasoning, they resolve the unknown attributes by providing related information (when A talks about ``Amy’’ ``swimming’’, B mentions ``Rowan College’’). When there is a conflict between the names, A promptly negates ``Ryan’’. 

\begin{table}[h!]
    \centering
    \resizebox{\linewidth}{!}{
    \begin{tabular}{l|cccc}
        \toprule
        \multicolumn{5}{c}{\textbf{Mutual Friend: alignment}}\\
        \toprule
        Groups & Success & Cooperative & Informative & Enjoyment \\
        \midrule
        GPT-3.5 w/o mind & 57.14 & 8.57 & 9.43 & 5.29 \\
        GPT-3.5 w/ mind & 62.50 & 8.88 & 9.63 & 7.63 \\
        \midrule
        \toprule
        \multicolumn{5}{c}{\textbf{CaSiNo: negotiation}}\\
        \toprule
        Groups & Scores & Skillful & Satisfied & Enjoyment \\
        \midrule
        GPT-3.5 w/o mind & 22.50 & 6.25 & 6.50 & 5.75 \\
        GPT-3.5 w/ mind & 24.50 & 7.13 & 7.25 & 7.25 \\
        \bottomrule
        \end{tabular}}
    \small
        \captionof{table}{\textbf{Human study}. Comparisons are made between our model with mind module vs. models w/o mind module when played with human subjects. }
    \label{tab:human}
\end{table}

\section{Conclusion}
In this study, we present MindDial, a novel framework for generating situated dialogue responses for common ground alignment and negotiation. 
By incorporating the first- and second-order ToM modeling into account, our model 
can enhance the alignment accuracy and negotiation outcome in both finetuning and prompting-based models. The efficacy of our approach is further substantiated through ablation studies and user feedback.

\section*{Limitations} Our prompting design for the mind modules requires a well-defined knowledge and goal. This may limit the generalization abilities of the current framework to more casual conversation scenarios. Also, the task success is highly dependent on the belief estimation precision. Future research is needed to develop and implement mind modules that are both more robust and accurate.

\section*{Ethics Statement}
Our work strictly relies on publicly available benchmarks and data, ensuring no personal information is collected. Our human study is conducted with utmost care to safeguard participants' privacy and interests, with no potential harm involved.

\bibliography{anthology,custom}

\appendix

\section{MutualFriend belief annotation and evaluation}
\label{sec:belief label}
\begin{figure*}[t!]
    \centering
    \small
    \includegraphics[width=\linewidth]{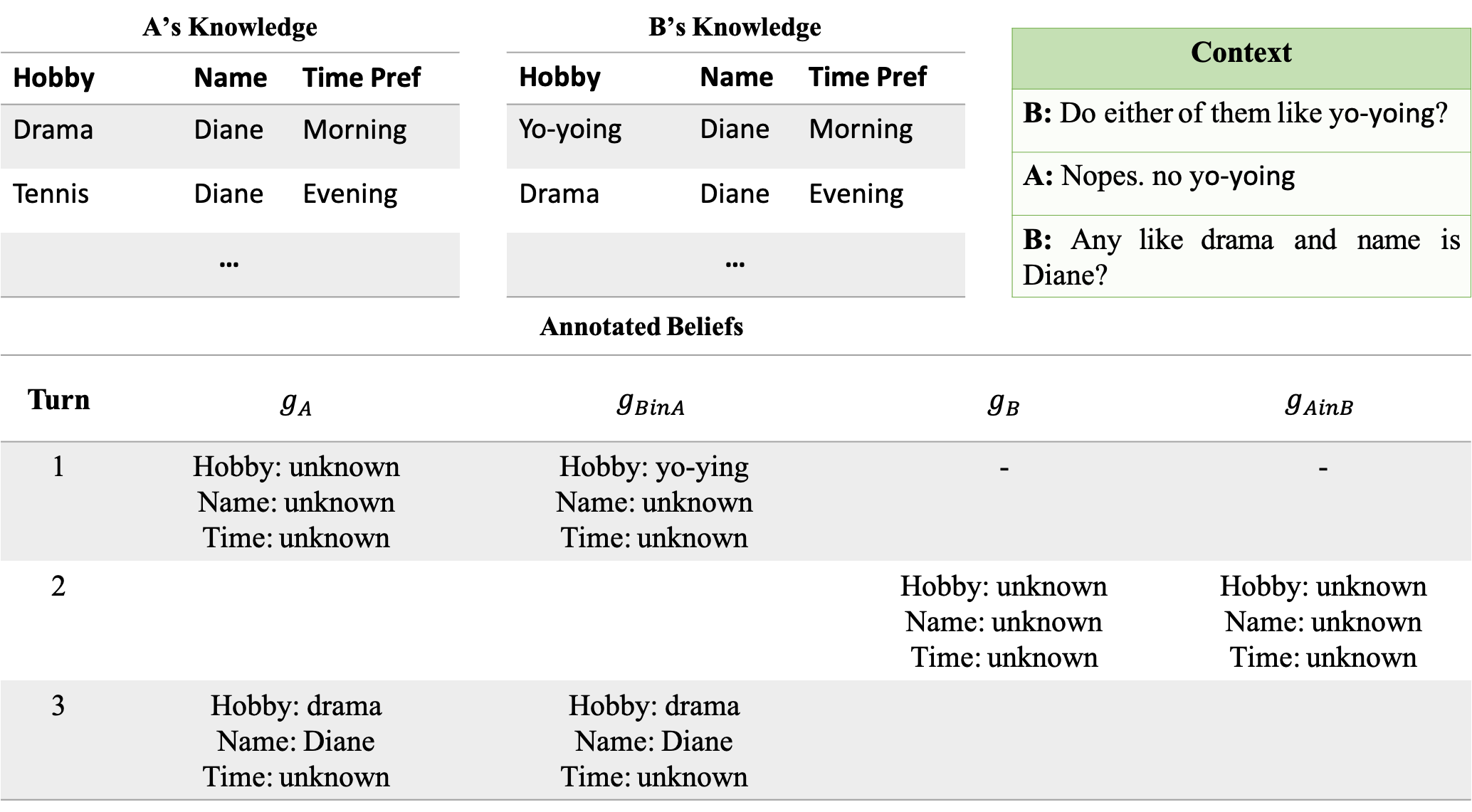}
    \caption{Annotation example}
    \label{fig:annt_example}
\end{figure*}

To test the belief estimation accuracy of our mind modules, we manually label the first and second-order beliefs given the current context of the dialogues. The values mentioned in the current dialogue context are marked as positive (1). The values not mentioned or negated by either of the agents are marked as negative (0). When all the values of one attribute are marked as negative, this attribute becomes ``unknown''. \Cref{fig:annt_example} illustrate one annotation process. For example, when B is asking about ``yo-yoing'', this value is marked as $1$ for $b_{BinA}$ hobby. However, since it does not belong to A's knowledge, for the first-order belief of speaker A, we annotate it as 0. Then, when ``yo-yoing'' is negated by A, it will be marked as 0 in $b_{BinA}$. The prediction is a true positive when the model's predicted value of one attribute is annotated as 1, a true negative when both prediction and ground truth are ``unknown''.

\section{MutualFriend prompts}\label{sec:mf_prompt}
\begin{center}
\begin{itemize}[leftmargin=*,itemsep=3mm,topsep=0pt]
    \item[$\rightarrow$] (At the beginning of the first turn): You are a smart cooperative agent named [Alice|Bob]. You have many friends with different attributes as listed below (the knowledge base of [Alice|Bob]). You are now talking with Bob. He also has a list of friends. You will talk with Bob for a maximum of 20 turns to find out your mutual friend as quickly as possible. You can ask him questions or provide information about your friends. Meanwhile, you should try to mention as few attributes and friends as possible.

hobby, name, location

Surfing, Jane, Outdoor

...

(After each turn - no mind):
\item [$\rightarrow$] [Alice|Bob] said: \{last generated response\}. Please provide your next utterance to [Alice|Bob]:
\item [$\rightarrow$]  Have you found your mutual friend? If yes, provide this mutual friend in the format of hobby|name|location; If no, respond 'unknown':

(After each turn - with mind):
\item [$\rightarrow$]  (first-order) Based on the current conversation and your friend table, who do you believe is your mutual friend? Respond in the format of hobby|name|location, and put unknown in the attributes you are not sure about for now: 

\item [$\rightarrow$]  (second-order) Based on the current conversation and your friend table, who do you believe that [Alice|Bob] believes your mutual friend is? Respond in the format of hobby|name|location, and put unknown in the attributes you are not sure about for now:

\item [$\rightarrow$] [Alice|Bob] said: \{last generated response\}. I estimate the mutual friend estimation from your perspective: [first-order] and from [Alice|Bob]'s perspective: [second-order] based on your current talk.  To align your estimation and resolve unknown attributes, please provide your next utterance to [Alice|Bob]:
\item [$\rightarrow$]  Have you found your mutual friend? If yes, provide this mutual friend in the format of hobby|name|location; If no, respond 'unknown':
\end{itemize}


\captionof{figure}{Template for MutualFriend self-talk prompt.}
\end{center}
\section{CaSiNo prompts}\label{sec:casino_prompt}
\begin{center}
\begin{itemize}[leftmargin=*,itemsep=3mm,topsep=0pt]
    \item[$\rightarrow$] (At the beginning of the first turn): You are a smart negotiation agent named [Alice|Bob] planning a camping trip. Besides basic supplies, you will need extra water, food, and firewood. Each of these items will be of either High, Medium, or Low priority for you as shown below. Each of them only has an available quantity of 3 and can only be split using integers. You will negotiate with Bob who will also need these items and have his own value table. Use reasons from your value table to justify why you need these items. Try hard to get as many items as you can!

Item, value, reason

water, high, I didn't pack enough water

...

(After each turn - no mind):
\item [$\rightarrow$] [Alice|Bob] said: \{last generated response\}. Please provide your next utterance to [Alice|Bob]:
\item [$\rightarrow$]  Based on your conversation with [Alice|Bob], do you want to end the negotiation? Please respond by yes or No:

(After each turn - with mind):
\item [$\rightarrow$]  (first-order) Based on the current conversation and your value table, how will you split water, firewood, and food? The items each person gets can only be integers and the total quantity for each item is 3. Please use the following format to respond without further explanation: item: the number you get/the number [Alice|Bob] get. For example, water:0/3, firewood:1/2, food: 3/0. 

\item [$\rightarrow$]  (second-order) Based on the current conversation and your value table, how do you think [Alice|Bob] will split water, firewood, and food? The items each person gets can only be integers and the total quantity for each item is 3. Please use the following format to respond without further explanation: item: the number you get/the number [Alice|Bob] get. For example, water:0/3, firewood:1/2, food: 3/0.
\item [$\rightarrow$] [Alice|Bob] said: \{last generated response\}. I estimated the negotiation deal from your perspective: [first-order] and from Bob's perspective: [second-order] based on your current talk. To align your expected deals, please provide your next utterance to [Alice|Bob]:
\item [$\rightarrow$]  Based on your conversation with [Alice|Bob], do you want to end the negotiation? Please respond by yes or No: 

(After negotiation ends):
\item [$\rightarrow$] Please provide your proposed deal. The items each person gets can only be integers and the total quantity for each item is 3. Deal with fractions will be rejected. Please use the following format: item: the number you get/the number [Alice|Bob] get. For example, water:0/3, firewood:1/2, food: 3/0.
\item [$\rightarrow$] Given your current conversation and the deal proposed by [Alice|Bob]: [deal], will you accept the deal? Please respond by Accept or Reject:
\end{itemize}


\captionof{figure}{Template for CaSiNo self-talk prompt.}
\end{center}
\section{Finetuning data format}


Generate the next response of the dialog based on the given context and knowledge:

    (SPEAKER0 as the current speaker)
    
Estimated [mutual friend|negotiation deal]
    
    [SPEAKER0] [First-order belief]
    
    [SPEAKER1] [Second-order belief]

Knowledge:
    
    Friend table or value table
    
Dialogues:
    
    [SPEAKER0] ...
    
    [SPEAKER1] ...
    
--- response:


\captionof{figure}{Template for Finetuning}

\section{Qualitative examples}
\subsection{MutualFriend}
\begin{figure*}[t!]
    \centering
    \small
    \includegraphics[width=\linewidth]{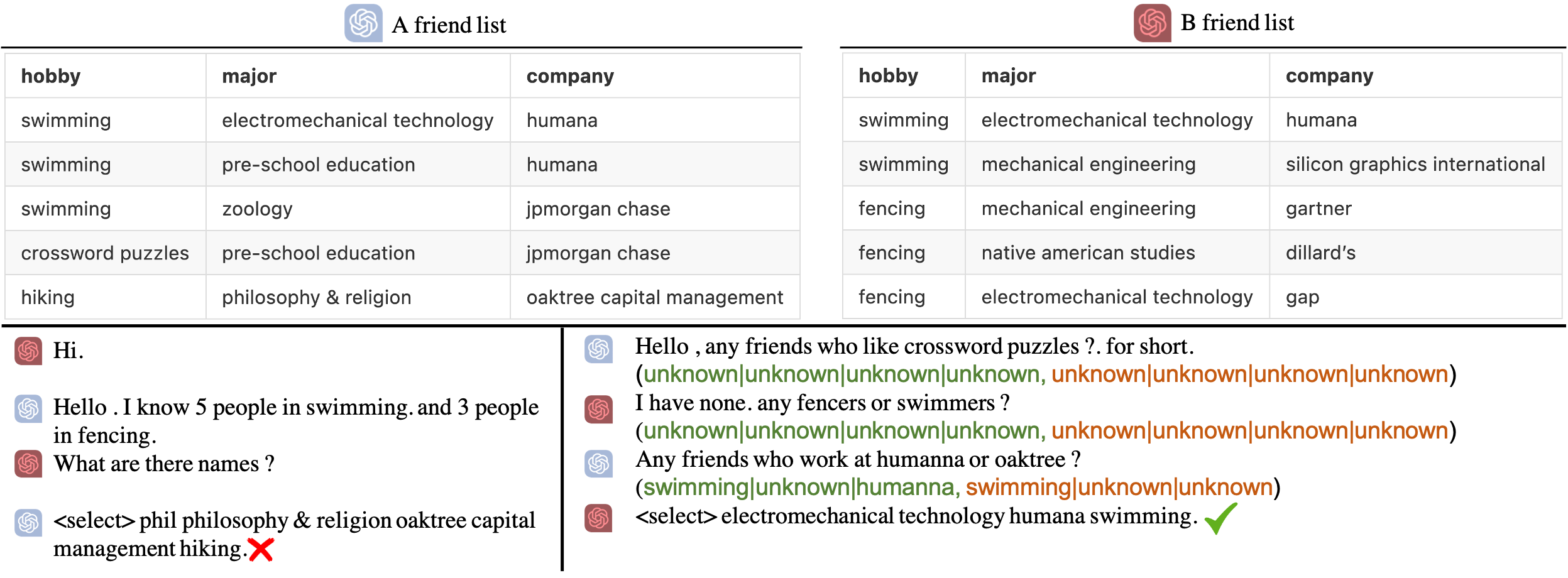}
    \caption{MutualFriend LLaMA-13B w/o mind vs mind}
    \label{fig:qual_mf_13b}
\end{figure*}

\begin{figure*}[t!]
    \centering
    \small
    \includegraphics[width=\linewidth]{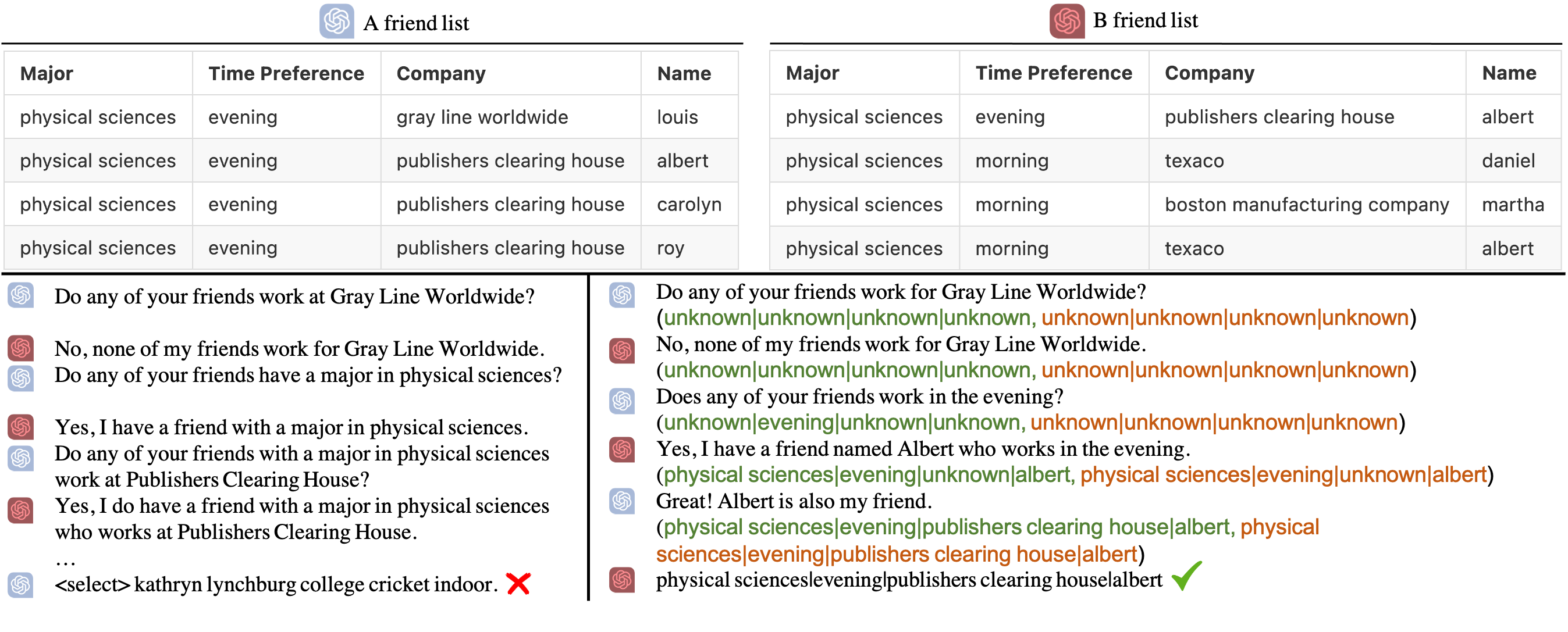}
    \caption{MutualFriend GPT-3.5 w/o mind vs mind}
    \label{fig:qual_mf_gpt3}
\end{figure*}

\begin{figure*}[t!]
    \centering
    \small
    \includegraphics[width=\linewidth]{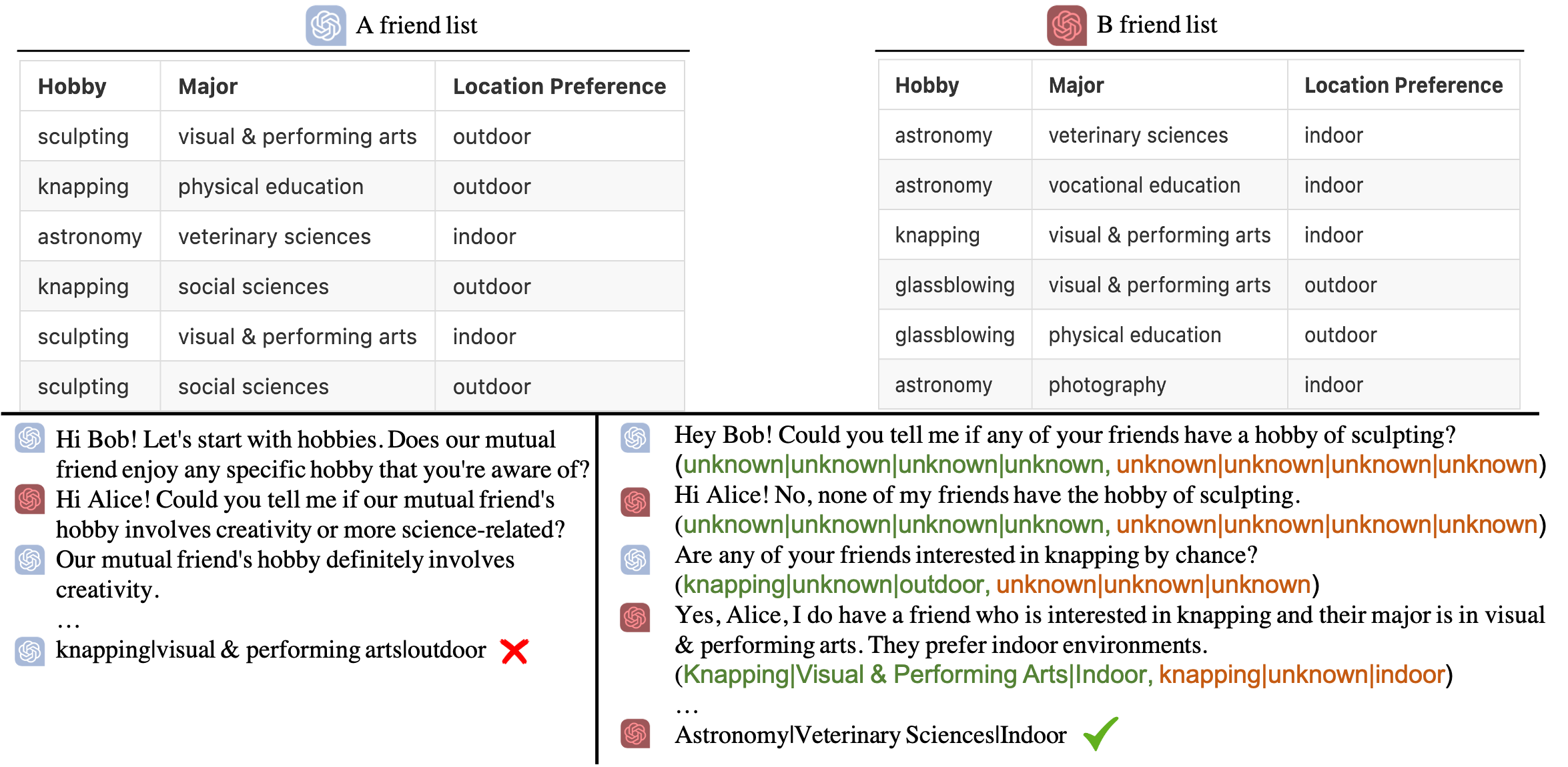}
    \caption{MutualFriend GPT-4 w/o mind vs mind}
    \label{fig:qual_mf_gpt4}
\end{figure*}

\cref{fig:qual_mf_13b}, \cref{fig:qual_mf_gpt3}, \cref{fig:qual_mf_gpt4} show the comparison between responses without and with mind reasoning in alignment settings. From~\cref{fig:qual_mf_13b}, we can see that LLaMA-13B without mind asks irrelevant questions like names. In~\cref{fig:qual_mf_gpt3}, GPT-3.5 augmented with the mind module can quickly merge to the common ground. For example, when A asks for ``evening'', B agrees and also provides ``Albert'' as relevant information. In contrast, the model without the mind module answers with the more rigid response patterns. GPT-4's performance without and with mind are comparable, but we find GPT-4 without mind modeling tends to talk longer and does not initiate with specific, targeted information.

\begin{table}[t!]
\centering
\small
\begin{tabular}{llccc}
    \toprule

    \textbf{Models} & \textbf{Sample size} & \textbf{$C$} & \textbf{$T$} & \textbf{$C_T$} \\
    \midrule
\multirow{4}{*}{LLaMA-13B-ft}&w/o mind & 36.33 & 6.64 & 5.47\\
& 1\% & 38.46 & 8.80 & 4.37 \\
&3\% & 44.67 & 8.85 & 5.05 \\
&5\% & 40.33 & 8.53 &  4.73\\
    \bottomrule
    \end{tabular}
\caption{\textbf{MutualFriend: results with different sample size}. }
\label{tab:mf_sample}
\end{table}

\begin{table}[th!]
\centering
\small
\begin{tabular}{llcc}
    \toprule

    \textbf{Models} & \textbf{Mind level} & \textbf{$C$} &  \textbf{$C_T$} \\
    \midrule
Human &- & 82.00 & 7.00\\
Rule & - & 90.00 & 5.00 \\
StanoNet &- & 78.00 & 4.00 \\
DynoNet &- & \textbf{96.00} & 6.00 \\
\midrule
LLaMA-7B-ft & w/o mind & 24.67 & 2.71 \\
LLaMA-7B-ft & $b_A+b_{BinA}$ & 28.33 & 3.20 \\
\midrule
LLaMA-13B-ft & w/o mind &36.33 & 5.47 \\
LLaMA-13B-ft & $b_A+b_{BinA}$ & 44.67 & 5.05 \\
\midrule
GPT-3.5 & w/o mind &10.67 & 1.86 \\
GPT-3.5 & $b_A+b_{BinA}$ & 24.33 & 4.03 \\
\midrule
GPT-4 & w/o mind &75.00 & 7.71 \\
GPT-4 & $b_A+b_{BinA}$ & 76.00 & \textbf{8.56} \\
    \bottomrule
    \end{tabular}
\caption{\textbf{MutualFriend: comparison with results from original paper}. }
\label{tab:mf_more_res}
\end{table}

\begin{table*}[t!]
\centering
\small
\resizebox{\textwidth}{!}{
\begin{tabular}{ll|cccccc}
    \toprule
    \textbf{Models} &\textbf{Sample size}& \textbf{Score-all} & \textbf{Sum} & \textbf{Agreed \%} & \textbf{Pareto} & \textbf{Score-agreed} & \textbf{Sum}\\
    \midrule
\multirow{4}{*}{LLaMA-13B-ft}&w/o mind & 15.38 vs 12.68 & 28.06 & 70.00 & 24.00 & 19.83 vs 15.97 & 35.80 \\
& 1\% & 15.36 vs 15.50 & 30.86 & 80.00 & 30.00 & 18.28 vs 18.46 & 36.74 \\
&3\%& 17.36 vs 17.32 & 34.68 & 92.00 & 40.00 & 18.43 vs 18.39 & 36.82 \\
& 5\% & 16.44 vs 16.58 & 33.02 & 86.00 & 34.00 & 18.30 vs 18.47 & 36.77 \\
    \bottomrule
    \end{tabular}}
\caption{\textbf{CaSiNo: results with different sample sizes}. }
\label{tab:casino_sample}
\end{table*}

\subsection{CaSiNo}~\label{sec:qual_casino}
\begin{figure*}[t!]
    \centering
    \small
    \includegraphics[width=\linewidth]{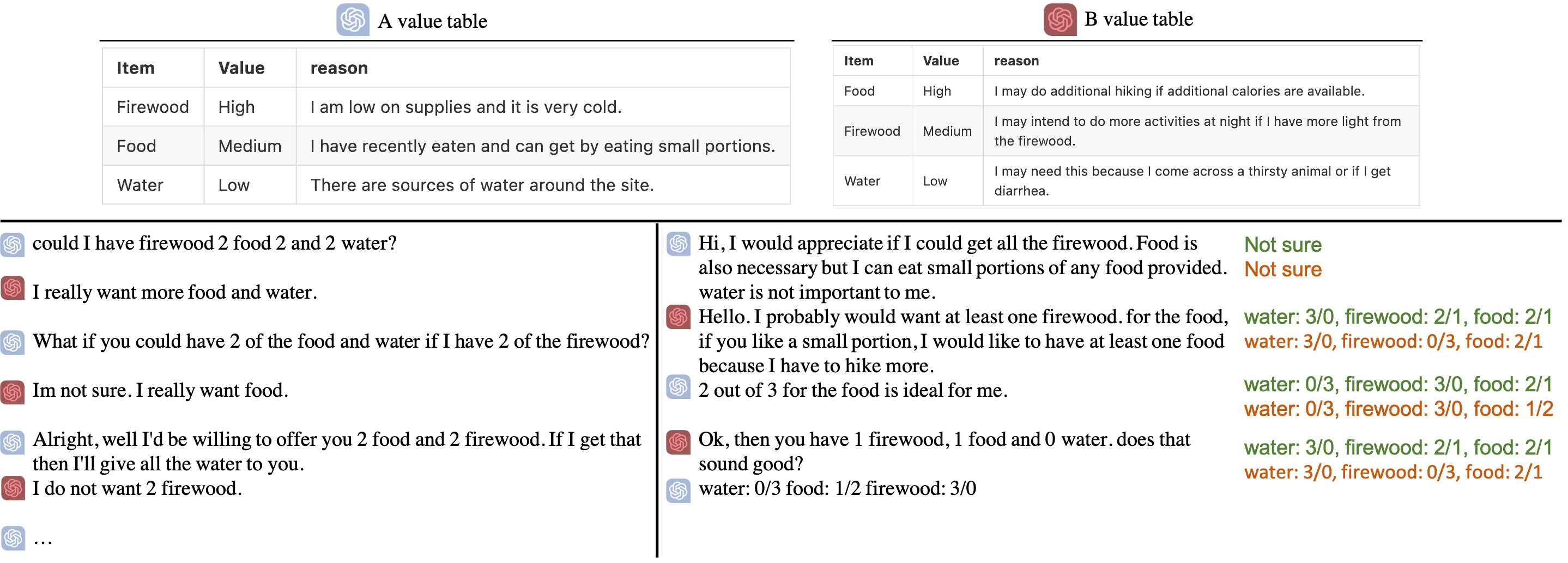}
    \caption{CaSiNo LLaMA-13B w/o mind vs mind}
    \label{fig:qual_casino_7b}
\end{figure*}

\begin{figure*}[t!]
    \centering
    \small
    \includegraphics[width=\linewidth]{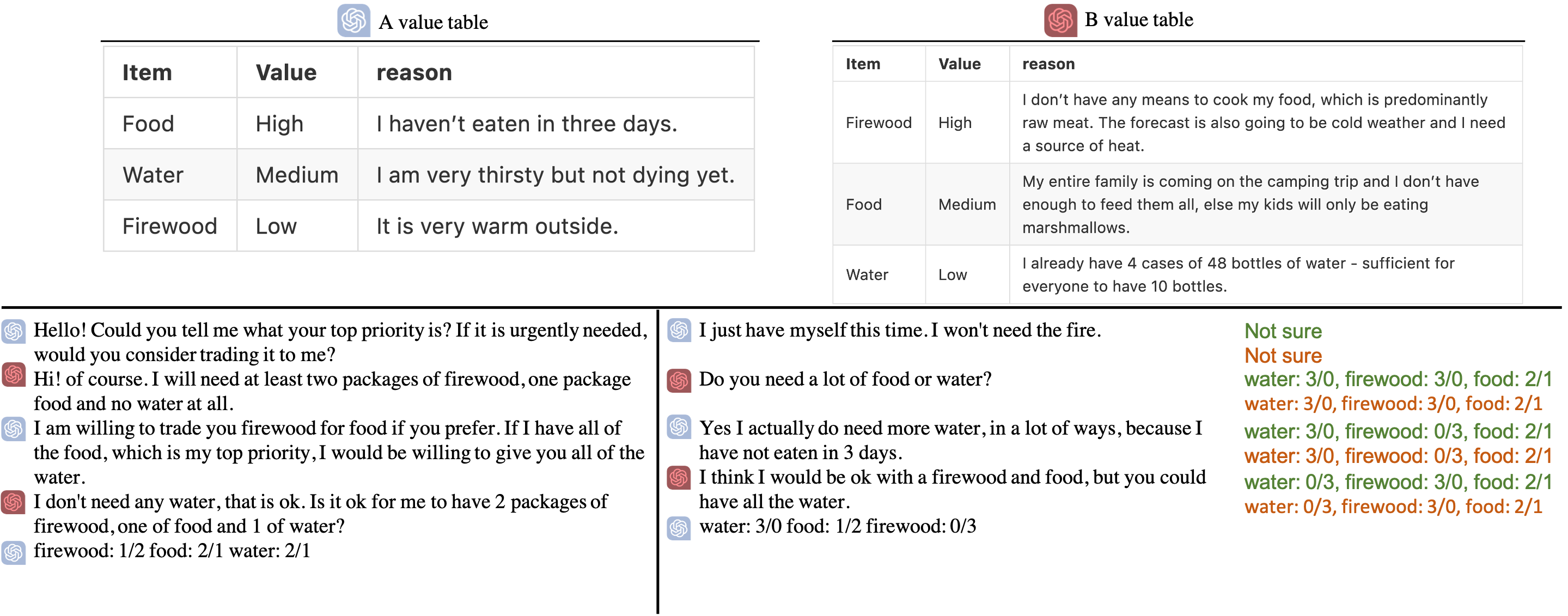}
    \caption{CaSiNo LLaMA-13B w/o mind vs mind}
    \label{fig:qual_casino_13b}
\end{figure*}

\begin{figure*}[t!]
    \centering
    \small
    \includegraphics[width=\linewidth]{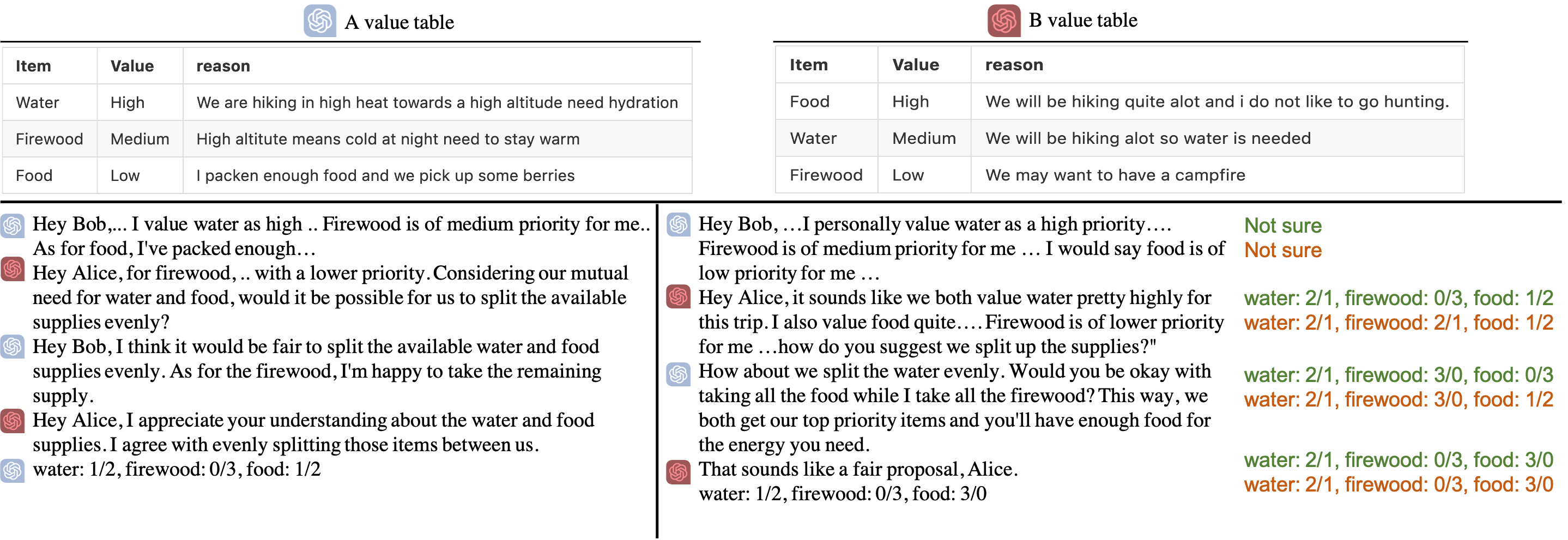}
    \caption{CaSiNo GPT-3.5 w/o mind vs mind}
    \label{fig:qual_casino_gpt3}
\end{figure*}

\begin{figure*}[t!]
    \centering
    \small
    \includegraphics[width=\linewidth]{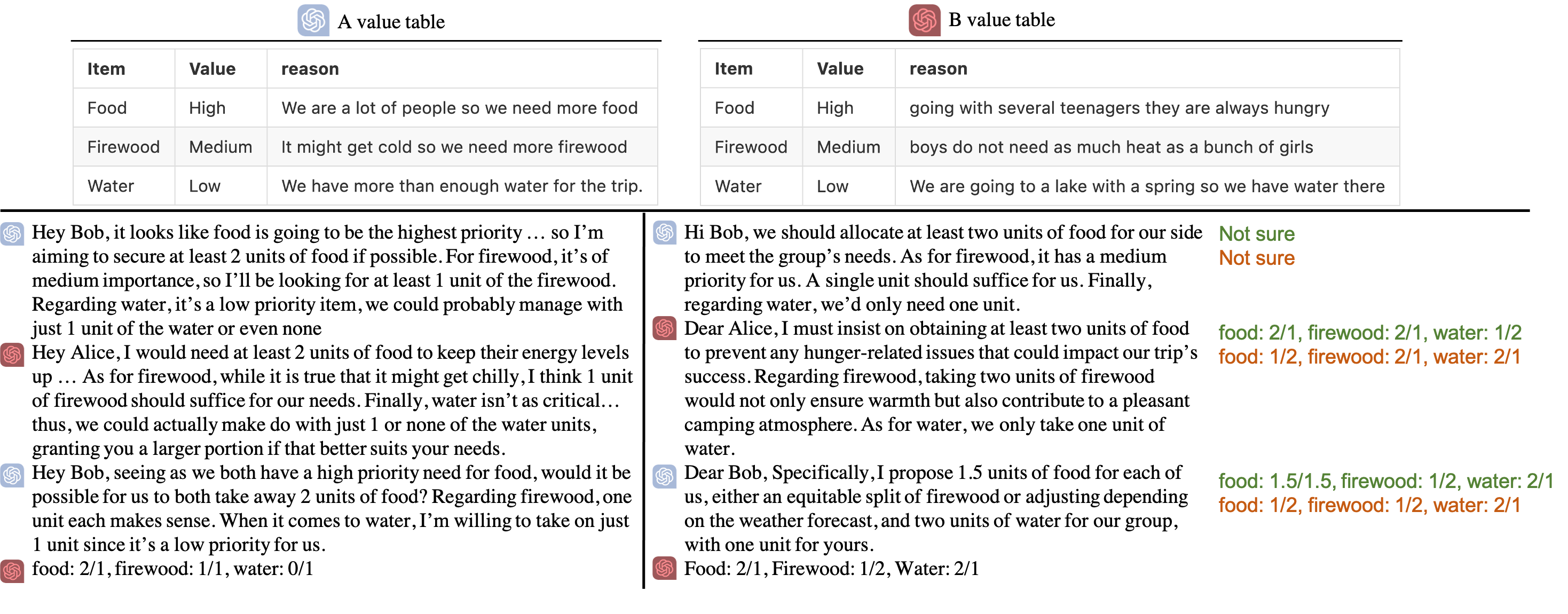}
    \caption{CaSiNo GPT-4 w/o mind vs mind}
    \label{fig:qual_casino_gpt4}
\end{figure*}

\cref{fig:qual_casino_7b}, \cref{fig:qual_casino_13b}, \cref{fig:qual_casino_gpt3}, \cref{fig:qual_casino_gpt4} show the comparison between responses without and with mind reasoning in negotiation settings. In~\cref{fig:qual_casino_7b}, we can see the model without mind reasoning cannot end the negotiation efficiently. From~\cref{fig:qual_casino_13b} and~\cref{fig:qual_casino_13b}, models with the mind can propose more Optimal distribution solutions and effectively address the most conflicted items. In~\cref{fig:qual_casino_gpt4}, we demonstrate an example when two agents have highly conflicted value table. Under this situation, GPT-4 leans towards a fair split. This strategy may cause the performance discrepancy we mention in~\cref{sec:quan}.

\section{Interface}\label{sec:interface}
\begin{figure*}[t!]
    \centering
    \small
    \includegraphics[width=\linewidth]{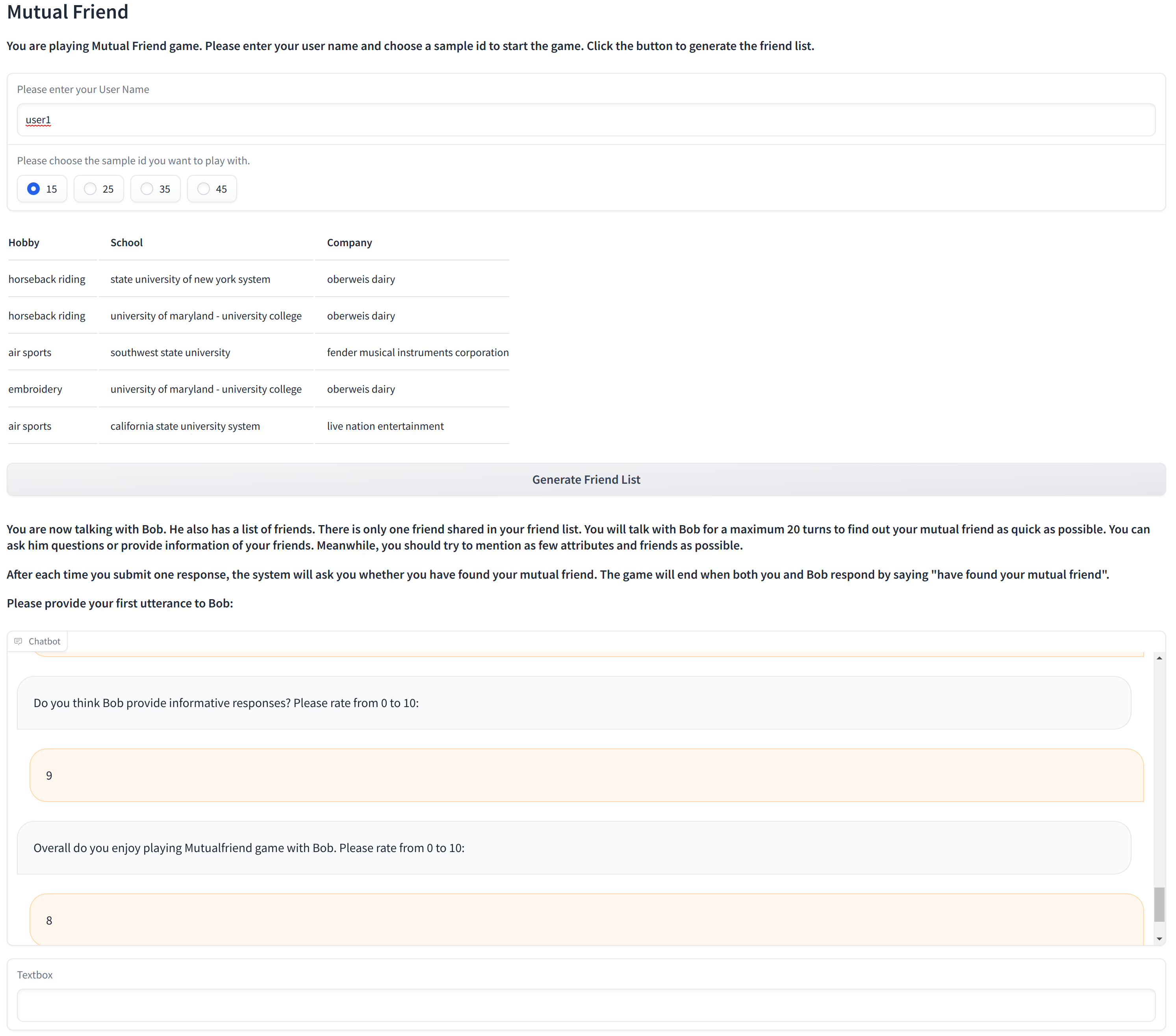}
    \caption{MutualFriend interface}
    \label{fig:interface_mf}
\end{figure*}

\begin{figure*}[t!]
    \centering
    \small
    \includegraphics[width=\linewidth]{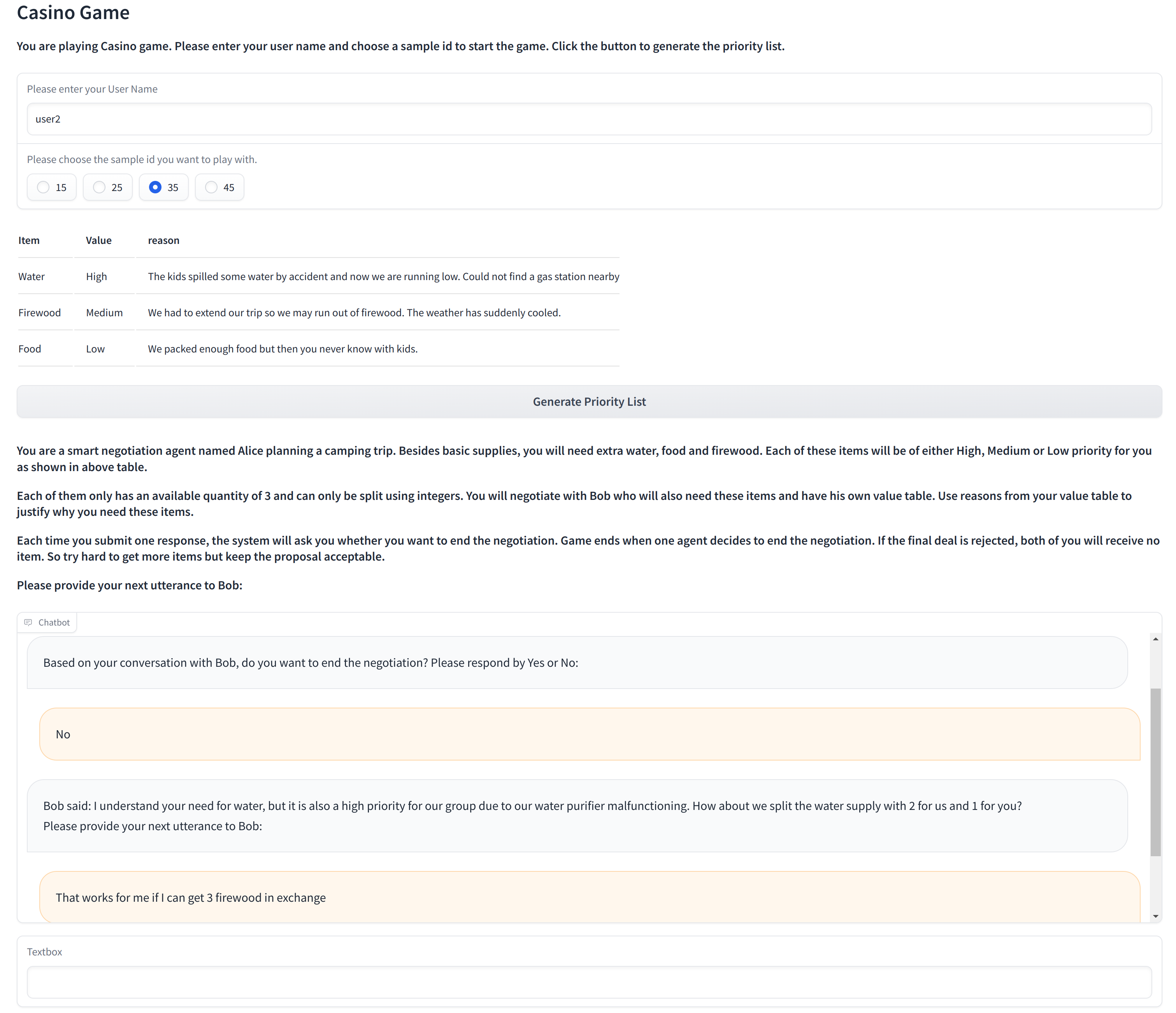}
    \caption{CaSiNo interface}
    \label{fig:interface_casino}
\end{figure*}

We build our interfaces using Gradio~\footnote{https://www.gradio.app/}. \cref{fig:interface_mf} is the example of the MutualFriend interface when the subject is giving ratings to the agent after the game ends. \cref{fig:interface_casino} shows one CaSiNo interface example when the subject is negotiating with the agent.
\section{Varing sample size of mind annotation data during finetuning}\label{sec:sample}
Considering the computational cost during fine-tuning, we only sample a small partition of dialogue for mind augmentation. In this section, we vary the sample size by 1\%, 3\% and 5\%. From~\cref{tab:mf_sample} and~\cref{tab:casino_sample}, we can see that 5\% achieves the best results and all models perform better than the w/o mind baselines.

\section{MutualFriend: more comparison results}\label{sec:more_mf_res}
In this section, we provide the baseline results of MutualFriend from the original paper in~\cref{tab:mf_more_res}. It is shown that GPT-4 can achieve higher efficiency with higher accuracy per turn. It is worth noting that the models in the original paper are of smaller sizes and trained with specific datasets while we currently focus more on larger models generalizable to more open-domain tasks. The CaSiNo dataset was originally designed for the strategy prediction task, therefore it did not report generation results. 

\end{document}